\documentclass{article}
\usepackage{arxiv}

\usepackage{geometry}
\usepackage{xcolor}
\usepackage{hyperref}
\usepackage{titlesec}
\usepackage{enumitem}
\usepackage{amsmath}
\usepackage{amssymb}
\usepackage{graphicx}
\usepackage{booktabs}
\usepackage{cite}
\usepackage{array}
\newcolumntype{C}[1]{>{\centering\arraybackslash}p{#1}}
\usepackage{xcolor}
\usepackage{colortbl}
\usepackage{comment}
\usepackage{subcaption}
\usepackage{makecell}
\usepackage{arydshln}
\usepackage{pifont}
\usepackage{xcolor}
\usepackage{booktabs}      % \toprule, \midrule, \bottomrule
\usepackage{multirow}      % for merged cells
\usepackage{siunitx}       % nice number alignment, optional
\usepackage{makecell}      % allow line‑breaks in header cells
\usepackage{amsmath}       % for \uparrow / \downarrow symbols
\usepackage{multirow}
\usepackage{siunitx}      % optional: nicer number alignment
\definecolor{alrow}{RGB}{220, 230, 242}

\definecolor{headerblue}{RGB}{220, 230, 242}
\definecolor{algreen}{RGB}{214, 234, 214}
\definecolor{improvegreen}{RGB}{0, 140, 0}
\definecolor{worsenred}{RGB}{180, 0, 0}
\definecolor{neutralgray}{RGB}{100, 100, 100}
\definecolor{bestrow}{RGB}{255, 243, 205}   % soft gold for 100% / best

\definecolor{feedsblue}{RGB}{30, 80, 160}

\title{%
  \textbf{Foundation Model-Enabled Efficient Data Sampling (FEEDS): A label-efficient training strategy for pan-cancer, multi-tracer PET/CT datasets}\\
}

\author{ 
	Biratal Raj Wagle \\
Department of Biomedical Data Science\\
Geisel School of Medicine at Dartmouth\\
Hanover, NH 03755, USA\\
	\texttt{Biratal.Raj.Wagle@dartmouth.edu} \\
    \And
    Bashirul Azam Biswas \\
Department of Biomedical Data Science\\
Geisel School of Medicine at Dartmouth\\
Hanover, NH 03755, USA\\
	\texttt{Bashirul.Azam.Biswas@dartmouth.edu} \\
	\And
	Grant Chau \\
Department of Biomedical Data Science\\
Geisel School of Medicine at Dartmouth\\
Hanover, NH 03755, USA\\
	\texttt{Grant.N.Chau.GR@dartmouth.edu} \\
    \And
	Matthew E. Maeder \\
Radiology\\
Dartmouth Hitchcock Medical Center\\
Lebanon, NH 03766 , USA\\
	\texttt{Matthew.E.Maeder@dartmouth.edu}\\ 
    \And
    Muhammad Azeem Arshad \\
Radiology\\
Dartmouth Hitchcock Medical Center\\
Lebanon, NH 03766 , USA\\
	\texttt{Azeem.Arshad@dartmouth.edu}\\ 
    \And
	Michael S. Leapman \\
Department of Urology\\
Yale University\\
New Haven, CT, USA\\
	\texttt{michael.leapman@yale.edu}\\ 
    \And
    James B. Yu \\
Radiation Oncology\\
Dartmouth Hitchcock Medical Center\\
Lebanon, NH 03766 , USA\\
	\texttt{James.B.Yu@dartmouth.edu}\\ 
    \And
	Indrani Bhattacharya \\
Department of Biomedical Data Science\\
Geisel School of Medicine at Dartmouth\\
Hanover, NH 03755, USA\\
	\texttt{Indrani.Bhattacharya@dartmouth.edu}\\
}
\date{}

\renewcommand{\shorttitle}{FEEDS}

\begin{document}

\maketitle

% ============================================================
\begin{abstract}
 Automated lesion segmentation in whole-body PET/CT imaging can assist clinicians with cancer detection, staging, and treatment planning across radiotracers and cancer types. However, training lesion segmentation models that capture variations in lesion size, distribution, and appearance requires large annotated datasets, whose creation is both time- and expertise-intensive. As a result, models trained on limited labeled PET/CT data often lack the accuracy and generalizability needed for clinical use. We present FEEDS (Foundation model-Enabled Efficient Data Sampling), a label- and compute-efficient learning strategy that uses vision foundation model embeddings to select the most informative and diverse unlabeled cases for expert annotation. Unlike unsupervised, semi-supervised, and active learning approaches, FEEDS is a one-step training paradigm requiring only a limited, representative training set, making it label- and compute-efficient. We train and validate FEEDS using the AutoPET-III dataset. We test its accuracy and generalizability on three held-out sets: AutoPET-III, DeepPSMA, and an internal Dartmouth-Hitchcock Medical Center dataset. We evaluate clinical utility at the voxel, lesion, and anatomic region level to assess performance in high-risk areas and treatment planning utility. FEEDS outperforms random-sampling-based labeling, pseudolabel-based semi-supervised learning, and training with limited labeled data alone. It generalizes across all three test sets, FDG and PSMA tracers, and multiple diseases, matching fully-labeled (100\%) training performance with 70\% less annotation burden. FEEDS addresses the challenge of label scarcity in an automatic lesion segmentation framework by providing a practical approach for constructing representative and diverse annotation queues from large, unannotated clinical repositories. Code is publicly available on \href{https://github.com/Image-and-Multimodal-Data-Analytics/FEEDS}{\color{blue}GitHub}   
\end{abstract}

\section{Introduction}
\label{sec:intro}

Whole-body lesion-segmentation on Positron Emission Tomography/Computed Tomography (PET/CT) is important for lesion detection, cancer staging, treatment planning, treatment response assessment, and outcome prediction~\cite{trotter2023positron}. However, interpretation is challenging due to nonspecific benign uptake, physiological variants, urinary tracer excretion obscuring bladder/urethral lesions, protocol and radiotracer differences, and subtle, heterogeneous uptake~\cite{hofman2016we,mingels2025total,belal2024applications}. Rising imaging volume, without a matching rise in trained nuclear medicine specialists, further burdens radiologists' workflows.

Deep learning-based lesion segmentation on whole-body PET/CT imaging can assist clinicians, but building robust, pan-cancer, multi-tracer models requires large annotated datasets, which are non-trivial to assemble. PET/CT lesion segmentation is not a part of routine clinical workflow, and radiology reports alone rarely yield strong segmentation labels. As the number of lesions, their size, distribution, and appearance vary widely by cancer type, disease spread, and imaging protocol, often with dozens of lesions per scan (Table \ref{tab:data_summary})\cite{gatidis2022whole,gatidis2024results,jeblick2026whole,meakin_2025_17701815}, exhaustive annotations is time- and expertise-intensive and adds to already overworked clinician workloads. Models trained on small datasets often fail to capture disease heterogeneity, limiting generalizability.

Existing approaches for limited labeled and large unlabeled datasets include unsupervised or self-supervised pretraining, semi-supervised learning (SSL), and active learning. Pretraining approaches learn representations from unlabeled data before fine-tuning, as in Yazdani \textit{et al.}~\cite{yazdani2024automated} (SwinUNETR on PSMA-PET/CT) and Patel \textit{et al.}~\cite{patel2022cross} (anomaly detection from healthy tissue). Foundation models like C²MAOT~\cite{huang2025c2maot} (cross-modal masked autoencoding for FDG PET/CT) and SegAnyPET~\cite{zhang2025seganypet} (promptable model trained on 5,731 PET volumes) have also shown strong generalization. SSL methods, broadly categorized into pseudolabeling, consistency regularization, GAN-based, contrastive, or hybrid~\cite{han2024deep} approaches, remain scarce for PET/CT and typically address a single tracer or cancer type, as in lymphoma~\cite{yousefirizi2024semi} and lung cancer~\cite{tang2026segmentation} and pan-cancer transfer learning~\cite{leung2024deep} studies. Leung \textit{et al.}~\cite{leung2024deep} is among the few studies addressing pan-cancer, multi-tracer settings, using complete and partial manual segmentations to iteratively refine lesions via semi-supervised transfer learning. Active learning iteratively selects informative cases for annotation to reduce labeling costs~\cite{biswas2023active}, typically via uncertainty estimation or latent-space clustering for diversity. Vali \textit{et al.}~\cite{vali2025active} fused both uncertainty estimation with latent-space feature clustering derived from self-supervised contrastive pretraining on the AutoPET-III dataset.

However, all these approaches require iterative training, demanding significant compute. No existing method achieves accurate, generalizable one-time training for pan-cancer, multi-tracer, whole-body PET/CT without heavy compute costs. Most methods are also single-tracer or disease-specific. Most studies rarely test generalizability on unseen datasets, and evaluate only via voxel-level metrics (e.g., Dice). Beyond voxel-level evaluation, lesion- and region-level assessments are crucial for identifying high-risk disease and guiding treatment.

To address these gaps, we propose FEEDS (Foundation model-Enabled Efficient Data Sampling), a label- and compute-efficient method that uses foundation model (DinoV2~\cite{oquab2023dinov2}) embeddings to select the most diverse, representative unlabeled cases for annotation before any model training. These cases are combined with the initial small labeled pool for one-time supervised training. With the AutoPET-III scans (training: 1043, validation: 247), our ablations identify how much representative data closes the gap to fully-labeled performance. Unlike active, unsupervised and semi-supervised learning methods requiring iterative or pretraining steps, FEEDS performs a single upfront case selection and a single training.

We evaluate FEEDS's accuracy and generalizability on three test sets: 321 AutoPET-III scans (FDG and PSMA; lung cancer, lymphoma, melanoma, prostate cancer, no cancer), 200 DEEP-PSMA scans (FDG and PSMA; prostate cancer), and 23 internal Dartmouth Hitchcock Medical Center scans (PSMA; prostate cancer). We include extensive evaluations to investigate the clinical utility of FEEDS in (a) voxel-level segmentation for lesion detection and extent, tumor-burden computation, and longitudinal tracking; (b) lesion counting for treatment planning and disease tracking; (c) anatomic region-level evaluation in high-risk areas relevant to treatment eligibility; and (d) performance consistency across disease and tracer types.

\section{Materials and Methods}

\subsection{Dataset}
\label{sec:data}

\textbf{Dataset:} Three retrospective, de-identified datasets are used: the publicly available Autopet-III \cite{Ingrisch2024_autoPETIII} and the Deep-PSMA \cite{meakin_2025_17701815} datasets, and an internal Dartmouth Hitchcock Medical Center (DHMC) dataset (Table \ref{tab:data_summary}). The AutoPET-III dataset includes $1,611$ whole-body PET/CT studies, with $1,014$ FDG-PET scans and $597$ PSMA-PET scans. The FDG PET scans include $513$ lesion-free, $188$ melanoma, $168$ lung cancer, and $145$ lymphoma cases. The PSMA-PET scans include $60$ lesion-free and $537$ prostate cancer patients. The Deep-PSMA dataset \cite{meakin_2025_17701815} consists of $100$ PET/CT patients with metastatic prostate cancer, each with an FDG and PSMA PET/CT scan acquired prior to LuPSMA therapy. The DHMC dataset includes $23$ PSMA-PET/CT scans from retrospective patients with or without prostate cancer who were seen at DHMC, Lebanon, NH. This retrospective study was approved by the Institutional Review Board (IRB) of Dartmouth College and Dartmouth Hitchcock Medical Center. As a chart review of previously collected data, patient consent was waived. 

For the AutoPET-III and DeepPSMA datasets, lesion segmentations were performed and/or verified by nuclear medicine specialists \cite{gatidis2022whole,jeblick2024psmapetctlesions}. For DHMC PSMA-PET/CT cases, candidate lesions were segmented by a medical image analysis researcher with one year of experience with PET/CT images (BAB) guided by nuclear medicine reports, then reviewed and refined by an expert Genitourinary Radiation Oncologist (JY). The number, location, intensity distribution of the lesions, and the scanning protocol in all three datasets vary significantly (Table \ref{tab:data_summary}), creating a diverse dataset. All images were preprocessed to remove intensity values below the  5th and above the 95th percentile followed by z-score normalization. 

 %In AutoPET-III, the number of lesions in FDG cases range from 1 to $1031$, whereas in PSMA-cases, it ranges from  1 to $294$. The PET intensity values also showed significant variability, with average intensity values within lesions ranging from $0.99$ to $31.06$ for FDG cases, and $0.74$ to $158.29$ for PSMA cases. Similarly, for DeepPSMA, the number of lesions varies considerably across patients, ranging from $1$ to $149$ in FDG cases and from $2$ to $235$ in PSMA cases. It may be noted that although each prostate cancer patient in DeepPSMA had both an FDG and PSMA scan, the number of visible lesions vary across the two tracers. The average PET intensity values within lesions range from $1.97$ to $19.38$ for FDG cases and from $2.99$ to $50.77$ for PSMA cases. {\color{red} In the DHMC dataset, number of lesions vary from ....., with intensity values ranging from...also include a sentence on scanners and update table to include dh and scanner infomation}. 

\textbf{Train/Val/Test Split:} We considered $1043$ pan-cancer, multi-tracer images from the AutoPET-III dataset as our entire training set. Of these, 103 scans (10\%, 66 FDG, 37 PSMA) were randomly selected as `fixed labeled data', with the remaining 940 cases forming an unlabeled pool. Our validation set includes $247$ cases from AutoPET-III. To prevent data leakage, splits ensured each patient's longitudinal scans appeared in only one of the train, validation, or test sets. The remaining $321$ scans from AutoPET-III, and the entire Deep-PSMA and DHMC datasets were considered held-out test sets.

\begin{table}[htbp]
\small
\centering
\caption{Dataset summary. M=Male, F=Female, NS=Not Specified, M=Melanoma, Lu=Lung Cancer, Ly=Lymphoma, Neg=Negative, Pr=Prostate, $dz$ = distance, $\mu_{PET} = $ Mean PET Intensity }
\label{tab:data_summary}
\begin{tabular}{@{}lccccc}
\hline
\multirow[]{2}{*}{Properties} & \multicolumn{2}{c}{AutoPET-III} & \multicolumn{2}{c}{Deep-PSMA} & DH \\
\cmidrule(r){2-3} \cmidrule(r){4-5}\cmidrule(r){6-6}
& FDG & PSMA & FDG & PSMA & PSMA\\ 
\hline
$\#$ Patients  & 1014 & 597 & 100 & 100 & 23 \\
Sex (M/F/NS) & 570 / 440/ 4 & 597/ 0 & 100/ 0 & 100/0 & 23/0 \\
%Age (years) & 59.4 ± 16.0 (11--95)  & 71.43 ± 8.18 (48--92) & N/A & N/A & --  \\
%Data  & PET-CT & PET-CT & PET-CT & PET-CT & PET-CT  \\ 
\hline
Disease dist. & {\begin{tabular}{c}  M/Lu/Ly/Neg  \\ 188/168/145/513 \end{tabular}}  & {\begin{tabular}{c} Pr/Neg\\ 537/60\end{tabular}} & {\begin{tabular}{c}Pr/Neg\\100/0\end{tabular}} &{\begin{tabular}{c} Pr/Neg\\100/0 \end{tabular}} & {\begin{tabular}{c} Pr/Neg\\23/0 \end{tabular}}\\
\hline
$\#$lesions (per case) & 15.15 $\pm$ 29.17 & 35.04 $\pm$ 51.70 & 38.38 $\pm$ 39.66 & 68.41 $\pm$ 58.35 & 10.13 $\pm$ 16.27 \\

$\mu_{PET}$ (per lesion) & 4.56 $\pm$ 2.32 & 7.48 $\pm$ 7.12 & 3.73 $\pm$ 1.19 & 5.64 ± 3.33 & 6.415 $\pm$ 5.182  \\

$\#$ slices (per vol) &  200--661 & 135--963 & 199--1206 & 195--1261 & 303--380 \\

In-plane res. (mm) & 2.04--2.04 & 2.73--4.07 & 2.73--5.47 & 1.59--5.47 & 0.98--1.52 \\

$dz$ bet. slices (mm) & 3.00--3.00 & 2.00--5.00 & 1.5--5.00 &  1.5--5.00 & 3.00-3.00\\
train/val/test & 664/151/199 & 379/96/122 &0/0/100 & 0/0/100 & 0/0/23 \\
\hline
\end{tabular}%
\end{table}
% ============================================================
\section{Methods}
\label{sec:methods}

\subsection{\textbf{F}oundation Model-\textbf{E}nabled \textbf{E}fficient \textbf{D}ata \textbf{S}ampling (FEEDS)}
\begin{figure}[htbp]
  \centering
  \includegraphics[width=\linewidth]{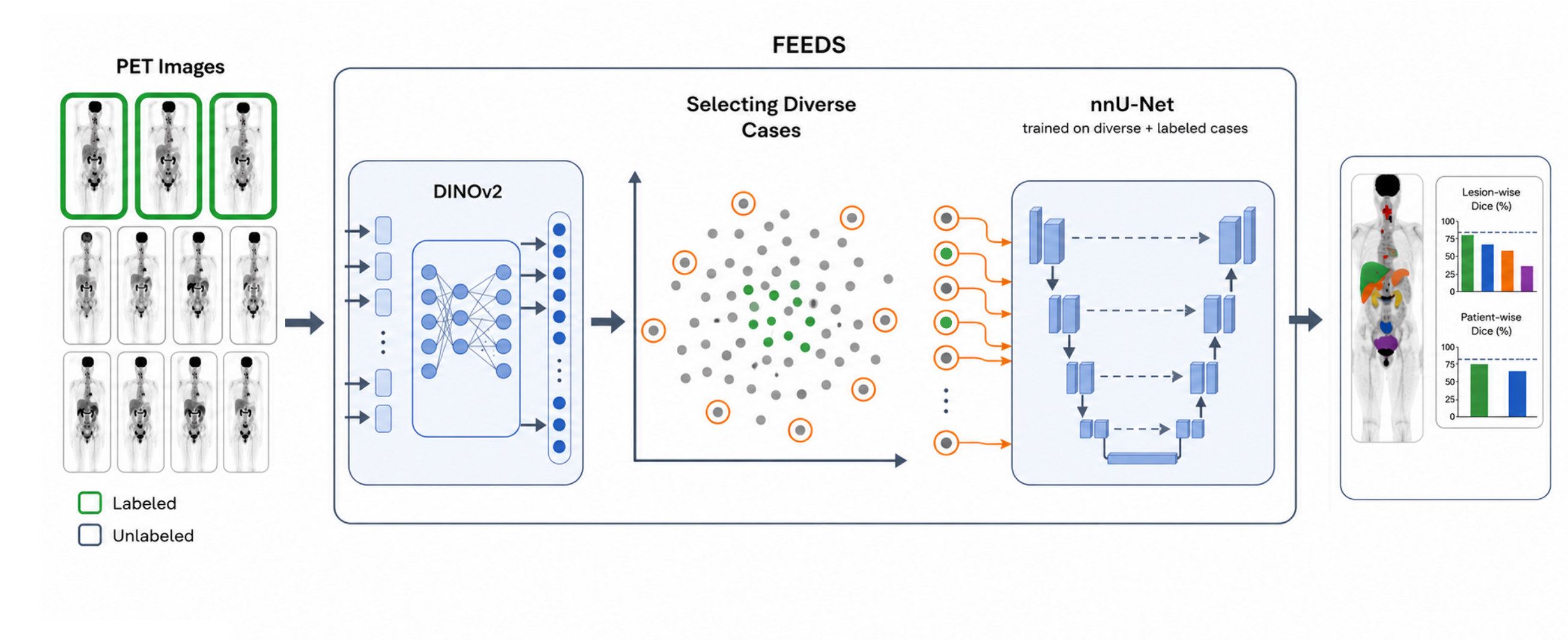}
  \caption{%
    Overview of the FEEDS: (1) foundation model (DinoV2) feature extraction from PET maximum intensity projection maps, (2) selection of unlabeled cases that are farthest from the labeled set in the feature space, (3) segmentation model training with combined labeled and newly annotated diverse samples.}
  \label{fig:feeds_overview}
\end{figure}

FEEDS achieves label- and compute-efficient whole-body lesion segmentation in three steps: (1) foundation model feature extraction, (2) diversity-based selection of unlabeled cases for annotation, and (3) segmentation model training on the combined labeled and newly annotated sets (Figure \ref{fig:feeds_overview}).

\subsubsection{Foundation model feature extraction}
For each 3D PET Image we compute  the maximum-intensity projection (MIP) images. 
The MIP images are z-score normalized, and passed through the pre-trained DinoV2 \cite{oquab2023dinov2} encoder to obtain image level token representations $f\in \mathbb{R}^{768}$. DinoV2 is a vision foundation model trained with 142 million natural images through self-supervision \cite{oquab2023dinov2}, and has been shown to learn general representations of medical images quite well \cite{baharoon2023evaluating, bhattacharya2025aggressiveness}. 

\subsubsection{Diversity-based selection of unlabeled cases for annotation}
\begin{figure} [!htbp]
    \centering
    \includegraphics[trim= 10 60 10 50, clip= true, width=\linewidth]{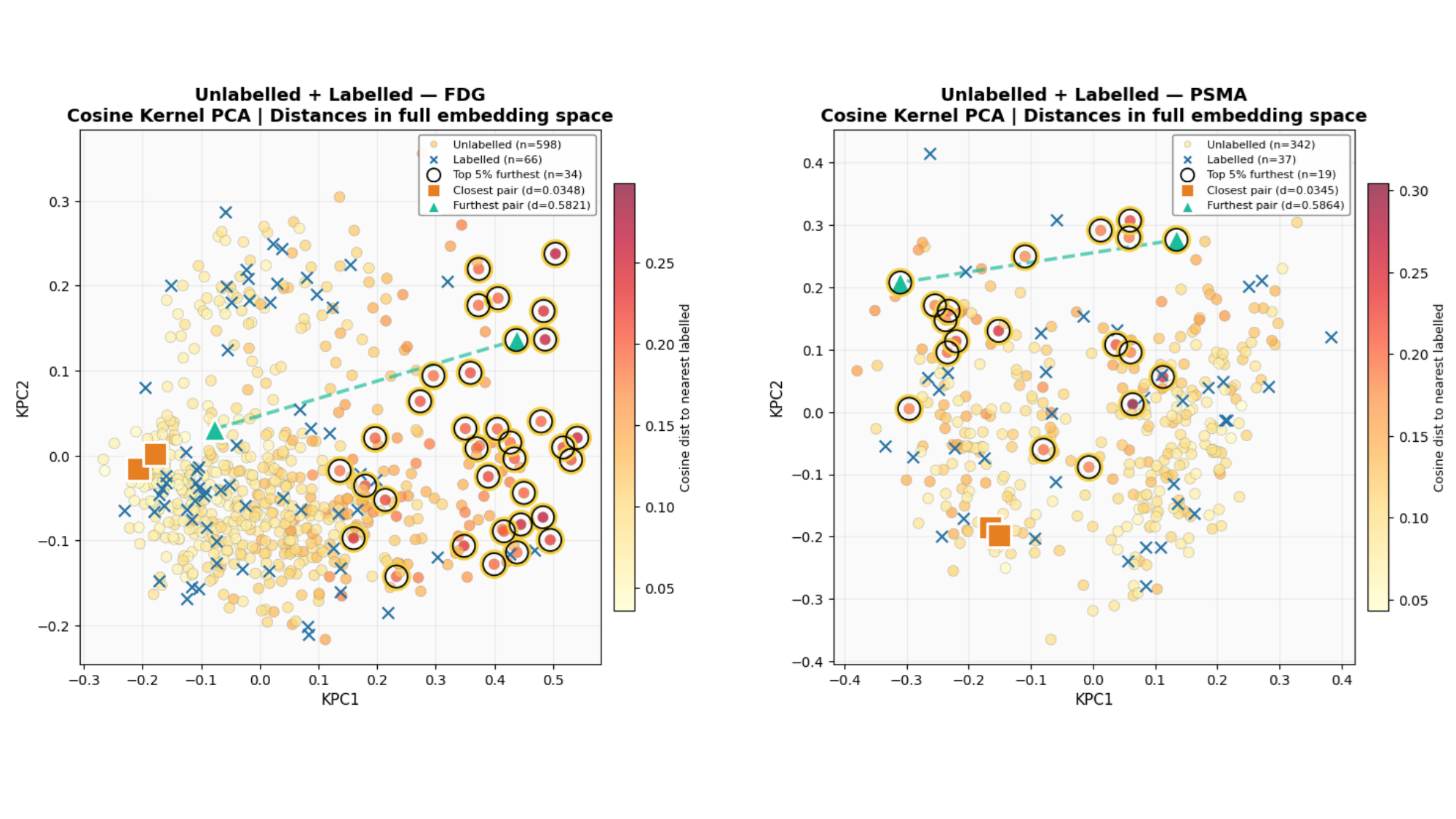}
    \caption{FEEDS samples additional cases for labeling by selecting the farthest cases (circled) from the labeled training cases (blue crossed) in the feature space. The two principal components of DinoV2 embeddings after Kernel PCA are shown here for visualization, but the $784$-dimensional feature space is used to compute distances and select cases. The sampling is done independently for FDG and PSMA tracers.}
    \label{fig:PCA_feeds}
\end{figure}

For each tracer-type $t$ (FDG or PSMA), let ${N_{L_t}}$ denote the small fixed labeled training data-pool, and ${N_{U_t}}$ denote the unlabeled training data pool. Let 
${L_{t}} = \{\mathbf{z}_i\}_{i=1}^{N_{L_t}}$ be the foundation model embeddings 
of the current labeled set for tracer-type $t$, and ${U_t} = \{\mathbf{z}_j\}_{j=1}^{N_{U_t}}$
be the embeddings of the unlabeled pool for the same tracer.
For each unlabeled tracer case $j$, we compute its minimum cosine distance
to the same tracer-specific labeled set:
\begin{equation}
  d_j = 1 - \max_{i \in{L_t}}
        \frac{\mathbf{z}_j \cdot \mathbf{z}_i}
             {\|\mathbf{z}_j\|\,\|\mathbf{z}_i\|}.
\end{equation}
Cases with the largest $d_j$ are those that are least represented with the fixed labeled training dataset. Since FDG and PSMA tracers show distinctly different clusters in feature space, we compute the distances based on tracer-type information. Within each tracer group, we select the farthest $X\%$ cases based on distance $d_j$. This promotes diversity and prioritizes underrepresented scan patterns, such as rare cancer types and unusual tracer uptake distributions, for labeling, while also preserving the FDG:PSMA ratio. We provide a visualization of the feature-distance-based sample selection process in Figure \ref{fig:PCA_feeds}.

\subsubsection{Segmentation model training on the combined labeled and newly annotated sets}
The FEEDS-sampled diverse cases are annotated and added to the small fixed pool of labeled cases and used to train an \textbf{nnU-Net}~\cite{isensee2021nnu} model for whole-body lesion segmentation. The nnU-Net model automatically configures preprocessing, patch size, network
topology, and post-processing based on dataset fingerprinting,
making it a robust and reproducible baseline for benchmarking
data selection strategies. The dataset is preprocessed as per Rokuss \textit{et al.}~\cite{rokuss2024fdg}, and the models are trained for 280 epochs.

\subsection{Ablation Experiments}

Using lesion segmentation performance on the  labeled validation set, our ablation experiments investigate:

\begin{enumerate}[label=(\alph*)]
\item \textbf{Effect of labeled training data size:} We monotonically increase labeled data by randomly sampling equal percentages of FDG and PSMA scans. We only perform random sampling once for each X\% increment.
\item \textbf{Effect of random sampling variability:} Starting from a fixed 10\% labeled dataset, we add X\% more cases to reach (10+X)\% labeled scans, repeating five random iterations per increment to generate different training sets at the same labeling budget.
\item \textbf{Effect of FEEDS-based selection:} For each X\% increase, we compare performance using FEEDS-selected versus randomly sampled cases.
\item \textbf{Effect of Determinantal Point Process (DPP)-based sampling:} Using the same foundation model features, we implement Determinantal Point Process (DPP)-based sampling and compare it against random and FEEDS-based sampling.
\item \textbf{Effect of iterative pseudo-label-based semi-supervised learning:} We first train an nnU-Net model on the fixed 10\% labeled pool and generate pseudolabels for the remaining 90\% unlabeled scans, then retrain using the 10\% strongly-labeled and 90\% pseudo-labeled data. We compare this semi-supervised approach against the other methods.
\end{enumerate}

% ============================================================
\subsection{Quantitative Evaluation Methods} 
\label{sec:metrics}

%We performed extensive clinically-relevant evaluations to assess how FEEDS performs in whole-body lesion segmentation on a (1) voxel-level, (2) lesion-level,  and (4) anatomical region-level. While voxel-level evaluation is typically used in medical image analysis tasks to assess segmentation performance via Dice, False Positive and False Negative Volumes, the lesion-, patient- and anatomical region-level evaluations are clinically-relevant in patient triage and treatment planning. We also assess performance in different diseases and across different tracers. We also assess generalizability of performance on three different datasets. To the best of our knowledge, there is no existing study to perform detailed clinically-relevant evaluations on pan-cancer, multi-tracer datasets. For each metric, numbers are computed for each patient, and average values over the entire validation or test set are presented.

\subsubsection{Voxel-level Evaluation}
Voxel-level evaluation assesses whole-body segmentation performance 
using Dice Coefficient, False Positive Volume (FPVol), and False Negative Volume (FNVol), similar to Rokuss \textit{et al.} \cite{rokuss2024fdg}. We present the aggregated evaluation over both tracer-types and all diseases, and also present stratified performance based on tracer- and disease-type. 

\begin{itemize}
  \item \textbf{Dice Similarity Coefficient (Dice, $\uparrow$):}
        Measures volumetric overlap between predicted and ground-truth
        lesion masks. Reported for diseased cases only.
  \item \textbf{False Positive Volume (FPVol, cubic cm (cc), $\downarrow$):}
        Total volume of spuriously detected tissue not present in
        the ground truth. Reported for all cases including
        non-diseased controls.
  \item \textbf{False Negative Volume (FNVol, cubic cm (cc), $\downarrow$):}
        Total volume of missed lesion tissue. Reported for diseased
        cases only (undefined for non-diseased).
\end{itemize}
\subsection{Lesion-Level Evaluation}

Lesion-level evaluation assesses how many lesions are correctly identified without regard to their full extent \cite{schott2025uncertainty}. Quantifying whole-body lesion count is important for treatment planning. For instance, in oligometastatic prostate cancer (disease spread beyond the pelvis but limited to 1–5 metastatic sites), aggressive metastasis-directed radiation therapy improves survival in this heterogeneous, poor-prognosis condition~\cite{alberto2022role, phillips2020outcomes}. Although the $\leq 5$-lesion oligometastatic threshold was derived from bone scans, accurate lesion counting remains critical for PSMA-PET–guided treatment decisions with radiation oncologists. For lesion-level evaluation, voxel-level ground truth labels were converted to discrete lesions via morphological processing and 3D connected component analysis. A lesion was considered detected if it overlapped with a prediction; ground truth lesions without any overlap were counted as false negatives. Lesions across all test patients were aggregated to compute lesion-level sensitivity and positive predictive value (PPV).

\subsection{Anatomical Region-Level Evaluation}
Treatment decisions for metastatic cancer often depend on disease presence in specific high-risk anatomical regions, and timely detection in these areas is critical for extending survival and quality of life. We evaluate model performance in four such regions: liver, lung, bone, and high-risk bones, motivated as follows:

\begin{itemize} \itemsep -2pt
 
    \item Our dataset spans lung cancer, prostate cancer, melanoma, and lymphoma.
    \item The liver is a common site of metastasis for many primary malignancies, causing significant cancer-related mortality and morbidity, occurring far more frequently than primary liver tumors \cite{ebrahim2025clinicopathological,tsilimigras2021liver, waninger2023evaluation, kadeerhan2023incidence}.
    \item Lung cancer is the most common cancer in the western world, with a 5-year survival of only 10-15\% \cite{beadsmoore2003classification}. 
    \item Bone metastases are a major source of cancer morbidity and mortality \cite{coleman2020bone}, causing pain, impaired mobility, pathological fracture, spinal cord compression, cranial nerve palsies, nerve root lesions, hypercalcemia, and bone marrow suppression. 
    \item Skeletal-related events (SREs), such as pathologic fractures and spinal cord compression, are a common consequence of bone metastasis. Radiation therapy for individual high-risk bone metastasis can decrease the occurrence of SREs, improve patient-centered outcomes (pain and subsequent hospitalization), and prolong overall survival. High-risk bone metastases is defined as fulfilling at least one of the following criteria: (a) Bulky site of disease in bone ($\geq$ 2 cm); (b) Disease involving the hip (acetabulum, femoral head, femoral neck), shoulder (acromion, glenoid, humeral head), or sacroiliac joints; (c) Disease in long bones occupying up to 2/3 of the cortical thickness (humerus, radius, ulna, clavicle, femur, tibia, fibula, metacarpals, phalanges); (d) Disease in junctional spine (C7-T1, T12-L1, L5-S1) and/or disease with posterolateral element (pedicles and/or facet joints) involvement \cite{gillespie2024prophylactic}.
\end{itemize}

We use the pre-trained TotalSegmentator model to segment the anatomical regions using CT scans. For each anatomic region of interest, we compute region-level sensitivity, specificity, Dice, FPVol, and FNVol. Region-level sensitivity and specificity assess whether the model detected at least one lesion in a high-risk region containing one or more true lesions; region-level Dice, false-positive, and false-negative volumes are computed from voxel-level metrics within each region. This analysis enables (a) identifying patients with region-specific disease spread for targeted treatment, and (b) assessing complete within-region disease extent to guide treatment planning (e.g., radiation or radioligand therapy).

% ============================================================
\section{Experimental Results}
\label{sec:results}

\subsection{Ablation Studies - Performance on AutoPet-III Validation Set}

We use voxel-level evaluation on the AutoPET-III validation set for our ablation experiments.

\begin{figure}[!htbp]
    \centering
    \includegraphics[width=0.32\linewidth]{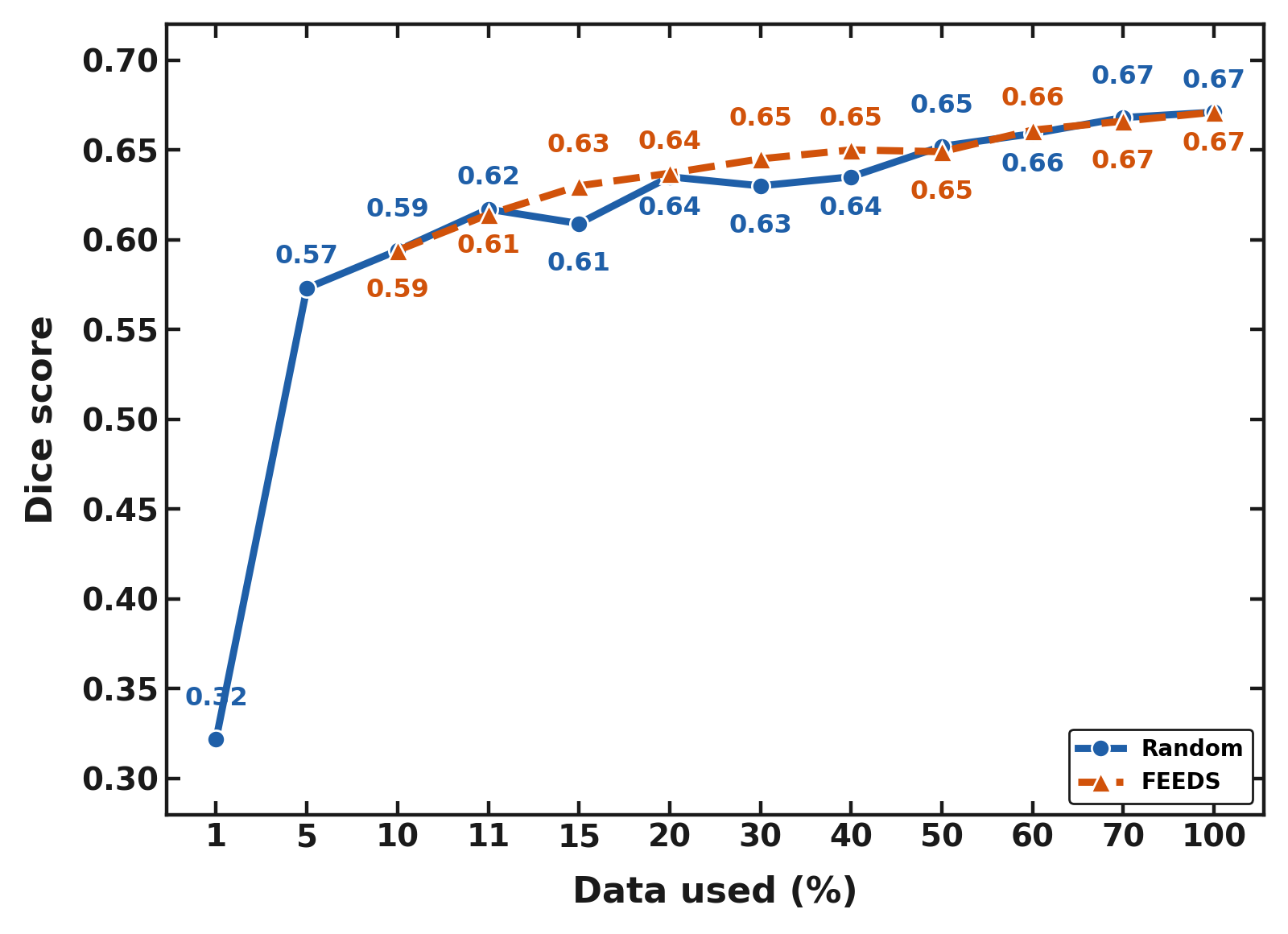}
     \includegraphics[width=0.32\linewidth]{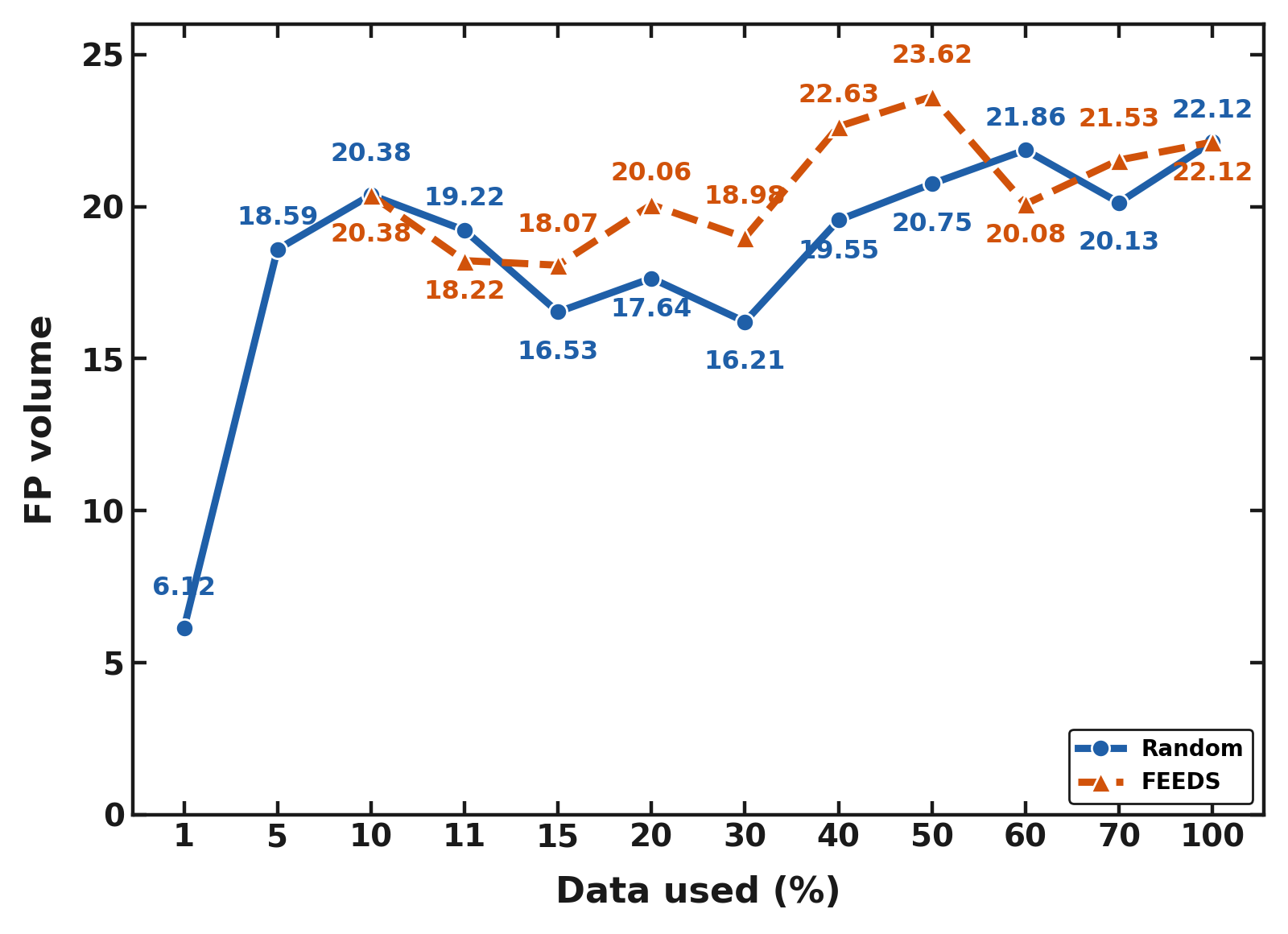}
     \includegraphics[width=0.32\linewidth]{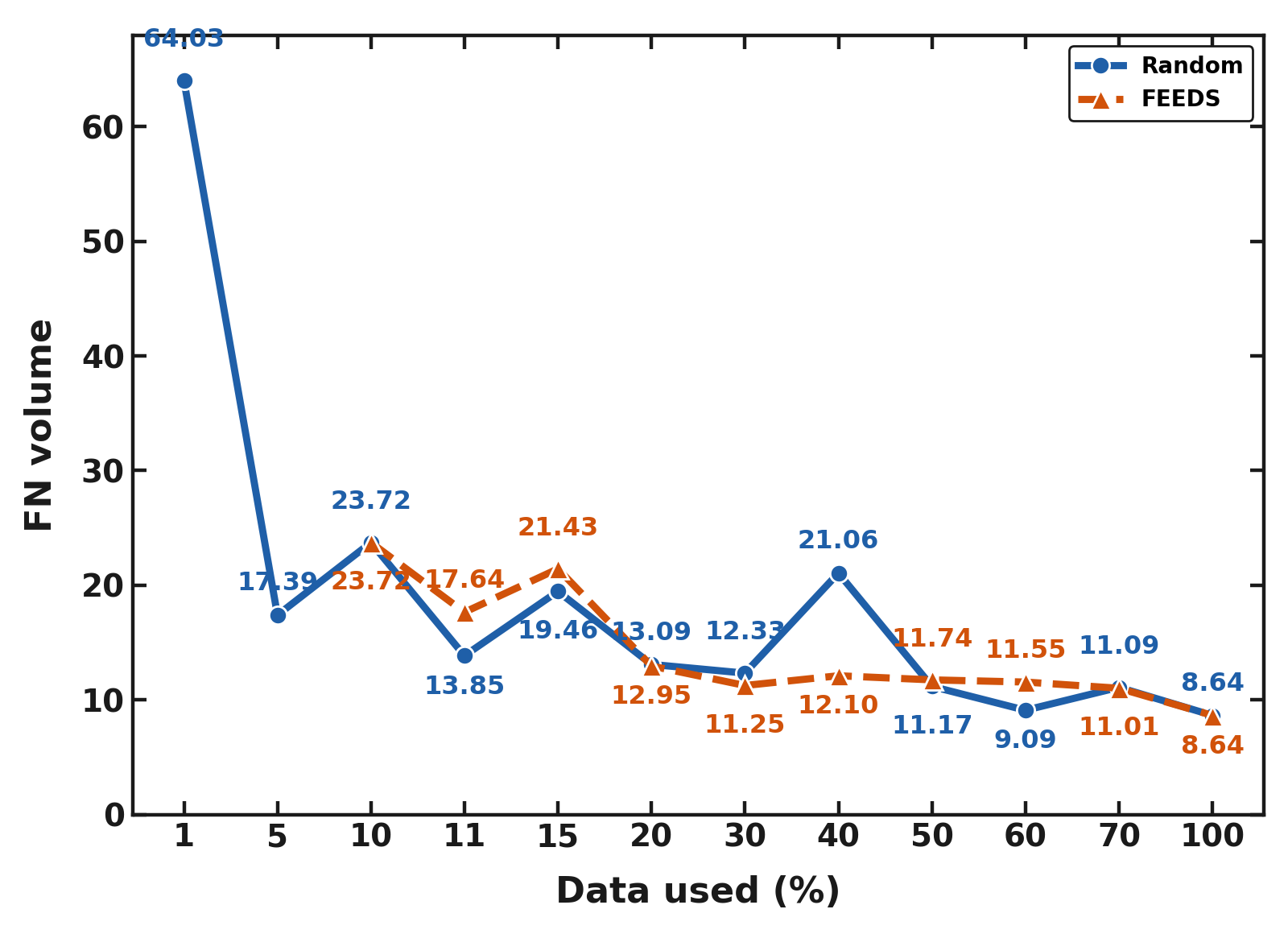}
    \caption{Whole-body PET/CT lesion segmentation performance improves with increasing percentage of labeled training data. FEEDS-sampled cases consistently improve Dice and false negative volumes compared to random sampling, where the performance can jump around.}
    \label{fig:random_vs_feeds}
\end{figure}

\textbf{Effect of labeled training data size (Figure \ref{fig:random_vs_feeds}):}
Segmentation performance improves with increasing labeled training data, with substantial gains between 1\% and 30\% and diminishing returns beyond. Since random sampling was performed only once per labeled data budget increase, performance varies non-monotonically due to the randomness in sampling. To assess variability across random sampling iterations, we repeated sampling five times at limited labeled data budgets (10\% fixed plus additional labeled data; Table~\ref{tab:ablation_full}). 

\begin{table}[!htbp]
    \centering
    \caption{Segmentation performance on the validation set (N=247) under varying labeled data budgets and sampling strategies. Our proposed FEEDS approach performs consistently better than randomly sampling cases for labeling, as well as the DPP sampling method. FDG total = 664 cases; PSMA total = 379 cases. Rand-1 through Rand-5 indicates iterations of random sampling; Rand-Mean, Rand-Std, and Rand-Median summarize these five iterations.
    $\uparrow$: higher is better; $\downarrow$: lower is better.}
    \label{tab:ablation_full}
    \renewcommand{\arraystretch}{1.2}
    \setlength{\tabcolsep}{6pt}
    \begin{tabular}{c c c c c c c}
        \toprule
        \textbf{Labelled Data Budget} & \textbf{FDG Cases} & \textbf{PSMA Cases} & \textbf{Sampling} & \textbf{Dice} & \textbf{FP Vol} & \textbf{FN Vol} \\
        \textbf{(total train cases)} & \textbf{(total=664)} & \textbf{(total=379)} & \textbf{Method} & ($\uparrow$) & ($\downarrow$) & ($\downarrow$) \\
        \midrule
        10\% (104)                       & 66        & 38        & Fixed          & 0.594 & 20.38 & 23.72 \\
        \midrule
        \multirow{10}{*}{10+10\% (208)}  & \multirow{10}{*}{66+66}  & \multirow{10}{*}{38+38}
            & Rand-1       & 0.635   & 17.64      &  13.09 \\
            &&& Rand-2       &  0.641   &  19.47 &  11.73 \\
            &&& Rand-3       & 0.638 & 16.73  & 12.26 \\
            &&& Rand-4       & 0.638 & 17.38 & 11.52 \\
            &&& Rand-5       & 0.634  &   19.59  &  23.83 \\
            \cdashline{4-7}
            &&& Rand-Mean    & 0.637 & 18.16 & 14.49 \\
            &&& Rand-Std     & 0.003 & 1.29  & 5.26  \\
            &&& Rand-Median  & 0.638 & 17.64 & 12.26 \\
            \cdashline{4-7}
            &&& DPP            & 0.622  & 18.78  & 22.62 \\
            &&& FEEDS          & 0.637  & 20.06 & 12.95 \\
        \midrule
        \multirow{10}{*}{10+20\% (313)}  & \multirow{10}{*}{66+133} & \multirow{10}{*}{38+76}
            & Rand-1       & 0.630 & 16.21 & 12.33 \\
            &&& Rand-2       & 0.641 & 19.03 & 13.03 \\
            &&& Rand-3       & 0.657 & 18.14  & 12.13 \\
            &&& Rand-4       & 0.636 & 20.70 & 13.44 \\
            &&& Rand-5       & 0.631  & 17.56 & 22.01 \\
            \cdashline{4-7}
            &&& Rand-Mean    & 0.639 & 18.33 & 14.59 \\
            &&& Rand-Std     & 0.011 & 1.68  & 4.18  \\
            &&& Rand-Median  & 0.636 & 18.14 & 13.03 \\
            \cdashline{4-7}
            &&& DPP            & 0.645 & 19.70 & 13.19 \\
            &&& FEEDS          & 0.645 & 18.98 & 11.25 \\
        \midrule
        \multirow{10}{*}{10+30\% (417)}  & \multirow{10}{*}{66+199} & \multirow{10}{*}{38+114}
            & Rand-1       & 0.635  & 19.55  & 21.06 \\
            &&& Rand-2       & 0.668 &      20.85   &     9.99 \\
            &&& Rand-3       & 0.655   &    18.17    &   17.39 \\
            &&& Rand-4       & 0.633   &    18.59     &  11.44 \\
            &&& Rand-5       & 0.648  &     18.09   &    12.07 \\
            \cdashline{4-7}
            &&& Rand-Mean    & 0.648 & 19.05 & 14.39 \\
            &&& Rand-Std     & 0.015 & 1.16  & 4.66  \\
            &&& Rand-Median  & 0.648 & 18.59 & 12.07 \\
            \cdashline{4-7}
            &&& DPP            & 0.656  &       17.87  &     17.67 \\
            &&& FEEDS          & 0.655 & 22.63 & 12.10 \\
        \midrule
        10+90\% pseudolabel (1043)       & 66+598    & 38+341    &       SSL      & 0.634 & -- & -- \\
        \midrule
        100\% (1043)                     & 664       & 379       & Fixed          & 0.671 & 22.12 & 8.64 \\
        \bottomrule
    \end{tabular}
\end{table}

\textbf{Effect of random sampling variability, FEEDS-based selection, DPP-based sampling, and semi-supervised learning (Table \ref{tab:ablation_full}):} 
Random sampling iterations showed wide variability even at the same labeled data budget (Table \ref{tab:ablation_full}). Averaged across five iterations, mean Dice scores were similar to or lower than FEEDS at 10+10\% (0.637 vs.~0.637), 10+20\% (0.639 vs.~0.645), and 10+30\% (0.648 vs.~0.655) budgets. Average false-positive volume was comparable or lower than FEEDS. The largest gains were observed in false-negative volume (FNVol), with random sampling yielding higher FNVol than FEEDS at every labeled data budget (14.49 vs.~12.95 at 10+10\%, 14.59 vs.~11.25 at 10+20\%, 14.39 vs.~12.10 at 10+30\%). Since FNVol only considers completely missed lesions, lower FNVol in FEEDS suggests more lesion detection. Random sampling variability is also reflected in standard deviations, highest for FNVol.

DPP with foundation model features performs comparably to FEEDS, though FEEDS achieves the lowest FNVol at each labeled data budget, suggesting greedy farthest-cases-first selection works well. Notably, FPVol is highest for the model trained on 100\% labeled data. Semi-supervised learning with weak pseudolabels (from a model trained on 10\% strongly-labeled data) performs worse than training on 10+10\% strongly-labeled data (Dice: 0.637 vs.~0.634), suggesting pseudolabels generated are noisy and can harm rather than help training.

Based on validation ablations, we selected the 10+20\% budget for test evaluation, using only 30\% of total labeled data. At this budget, FEEDS performed slightly below 100\%-labeled training but significantly above 10\%-labeled training on voxel-level evaluation.

\subsection{Held Out AutoPET-III Test Set Performance}

\subsubsection{Voxel-Level Evaluation}

\begin{figure}[htbp]
	\centering
	\begin{subfigure}[b]{1\textwidth}
	\centering
	\includegraphics[width=\textwidth]{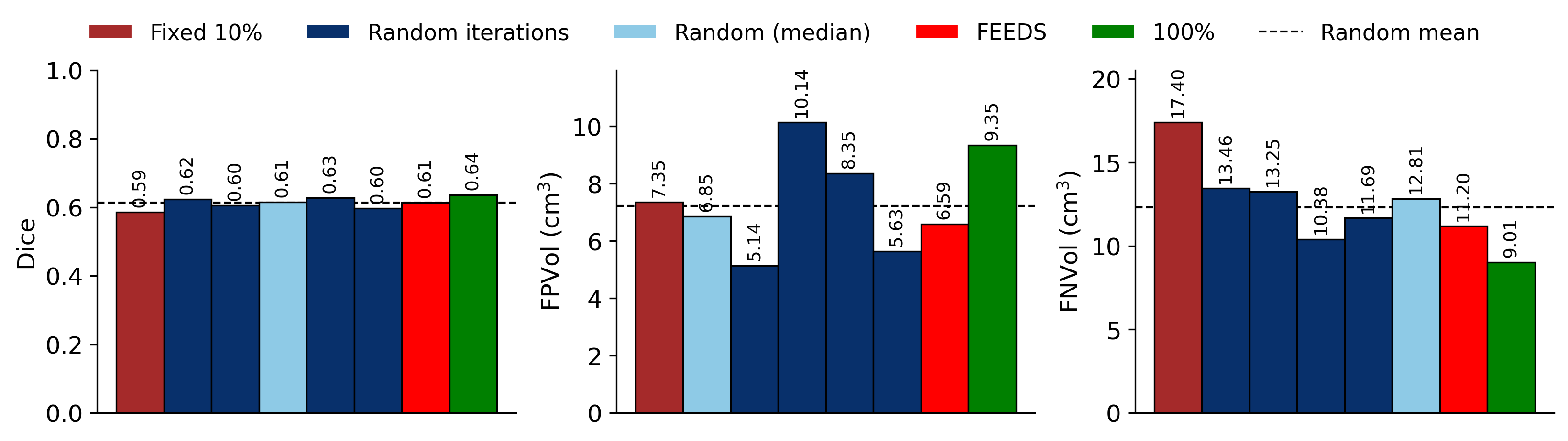}
	\caption{ ALL}
	\label{figs:ALL_test_random_vs_feeds}
	\end{subfigure}
	\begin{subfigure}[b]{1\textwidth}
		\centering
	\includegraphics[width=\textwidth]{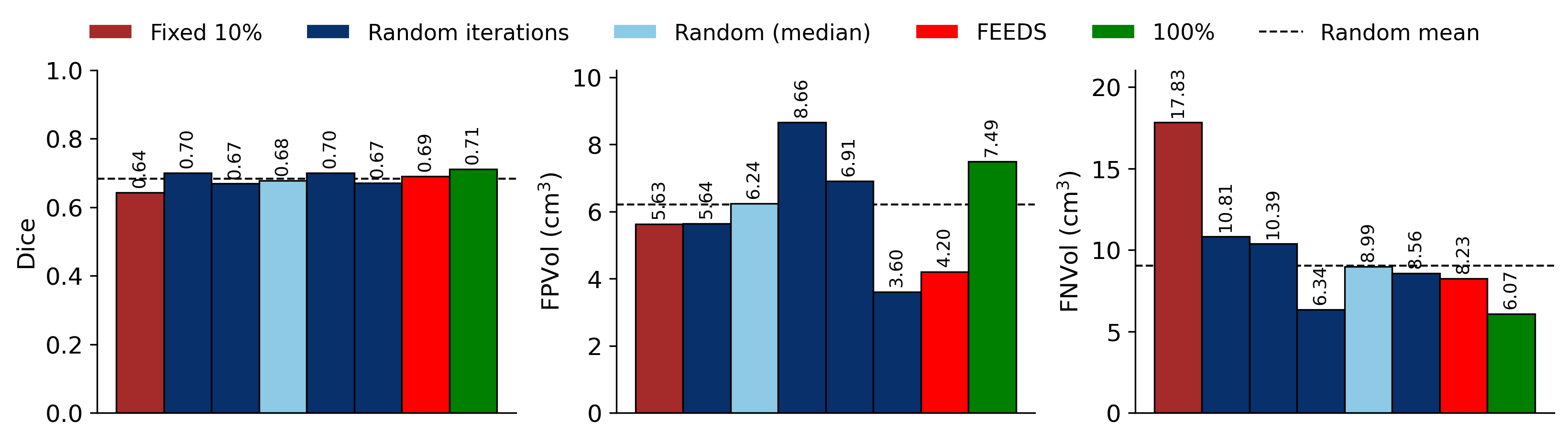}
	\caption{ FDG}
    \label{figs:FDG_test_random_vs_feeds}
	\end{subfigure}
    \begin{subfigure}[b]{1\textwidth}
		\centering
	\includegraphics[width=\textwidth]{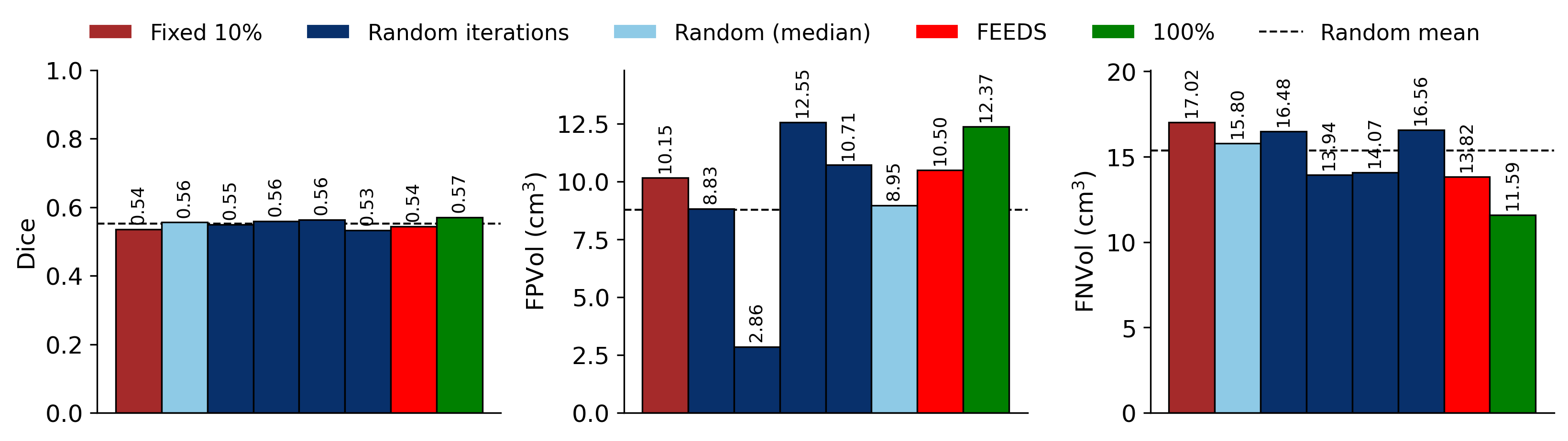}
	\caption{ PSMA}
    \label{figs:PSMA_test_random_vs_feeds}
	\end{subfigure}
    \caption[example]{Voxel-level evaluation on the AutoPET-III test set, comparing 10\% , (10+20)\% randomly sampled (five iterations), (10 + 20)\% FEEDS-selected, and 100\% labeled training cases. FEEDS outperforms 10\% and (10+20)\% random sampling, achieving near performance with 100\% labeled training set, in overall and tracer-specific evaluation.}
    \label{figs:summary_iterations_vs_feeds_test}
\end{figure}

\hspace{2pt} \textbf{Overall and tracer-stratified performance (Figure \ref{figs:summary_iterations_vs_feeds_test})} FEEDS outperforms random sampling-based training for the same labeled data budget across all metrics, except for PSMA FPVol. The model trained with 10+20\% FEEDS-sampled training cases achieve slightly lower Dice and FNVol to the model trained with 100\% labeled training cases (overall Dice: 0.61 vs.~0.64, FNVol: 11.20 vs.~9.01 cc) with better FPVol (6.59 vs.~9.35 cc). 

\begin{table}[!htbp]
    \centering
    \caption{Segmentation performance at the 10+20\% labeled data budget, broken down by disease type, compared against the 10\% fixed-labeled data baseline and the 100\% labeled training data upper bound. Rand-Mean and Rand-Std summarize five iterations of random sampling. FPVol and FNVol are reported in cubic cm.
    $\uparrow$: higher is better; $\downarrow$: lower is better. Within each disease type, \textbf{bold} indicates the best value and \underline{underline} the second-best value among Fixed 10\%, Rand-Mean, FEEDS, and 100\% (Rand-Std is excluded, as it is a variability measure rather than a performance score). Tied values receive the same mark.}
    \label{tab:disease_eval}
    \renewcommand{\arraystretch}{1.2}
    \setlength{\tabcolsep}{6pt}
    \begin{tabular}{c c c c c}
        \toprule
        \textbf{Disease Type} & \textbf{Sampling Method} & \textbf{Dice} & \textbf{FP Vol} & \textbf{FN Vol} \\
        & (10+20\% labeled data) & ($\uparrow$) & ($\downarrow$) & ($\downarrow$) \\
        \midrule
        \multirow{5}{*}{Lung Cancer}
            & Fixed 10\%   & 0.70 & 6.21 & 34.78 \\
            & Rand-Mean    & \underline{0.74} & \underline{3.30} & 15.01 \\
            & Rand-Std  & 0.017   &  1.828    & 1.954          \\
            \cdashline{2-5}
            & FEEDS        & 0.73 & \textbf{1.96} & \underline{14.14} \\
            & 100\%        & \textbf{0.75} & 4.05 & \textbf{9.97}  \\
        \midrule
        \multirow{5}{*}{Lymphoma}
            & Fixed 10\%   & 0.62 & 4.38 & 12.71 \\
            & Rand-Mean    & 0.66 & 4.33 & 8.41  \\
            & Rand-Std     &  0.031    &   1.195    & 2.938      \\
            \cdashline{2-5}
            & FEEDS        & \underline{0.69} & \textbf{2.15} & \underline{6.79}  \\
            & 100\%        & \textbf{0.73} & \underline{3.62} & \textbf{4.91}  \\
        \midrule
        \multirow{5}{*}{Melanoma}
            & Fixed 10\%   & 0.61 & \textbf{2.75} & 4.09 \\
            & Rand-Mean    & \underline{0.65} & 5.09 & \underline{3.04} \\
            & Rand-Std     &  0.012    &  1.391     &  0.701     \\
            \cdashline{2-5}
            & FEEDS        & \textbf{0.66} & \underline{4.07} & 3.13 \\
            & 100\%        & \underline{0.65} & 7.35 & \textbf{2.91} \\
        \midrule
        \multirow{5}{*}{Prostate Cancer}
            & Fixed 10\%   & 0.54 & \underline{10.15} & 17.02 \\
            & Rand-Mean    & \underline{0.55} & \textbf{8.78}  & 15.37 \\
            & Rand-Std     & 0.013     &  3.643    &  1.280     \\
            \cdashline{2-5}
            & FEEDS        & 0.54 & 10.50 & \underline{13.82} \\
            & 100\%        & \textbf{0.57} & 12.37 & \textbf{11.59} \\
        \midrule
        \multirow{5}{*}{Negative}
            & Fixed 10\%   & -- & 6.70 & -- \\
            & Rand-Mean    & -- &8.11 & -- \\
            & Rand-Std     & -- &2.52 & --\\
            \cdashline{2-5}
            & FEEDS        & -- & \textbf{5.61} & -- \\
            & 100\%        & -- & 9.85& --\\
        \bottomrule
    \end{tabular}
\end{table}

\textbf{Disease-stratified performance (Table \ref{tab:disease_eval}):} FEEDS with 30\% labeled training data budget achieves Dice comparable to 100\% labeled training data budget for all diseases. FNVol and FPVOl in FEEDS are lower than random sampling for most diseases. Notably, FEEDS reduces FPVOl in negative cases, indicating that diversity-driven selection limits over-segmentation in disease-free patients, a clinically important property. %In contrast, random sampling increases false positive volume relative to the 10\%-fixed baseline for melanoma and negative cases, while remaining unchanged for lymphoma. This disease-stratified evaluation shows that FEEDS selects diverse training samples across disease types, yielding performance gains across all disease types and negative cases.

\subsubsection{Lesion-Level Evaluation (Table \ref{tab:results_pat_lesion})}
FEEDS has higher lesion-level sensitivity than random sampling (0.743 vs.~0.728), and achieves the best PPV,  outperforming 100\% labeled training data (0.821 vs.~0.784). 
\begin{table}[!htbp]
\centering
\caption{Lesion-level evaluation comparing 10\% fixed labeled, (10+20)\% random and FEEDS-sampled labeled training cases, and 100\% labeled training cases. Lesion sensitivity and positive predictive value (PPV) are reported micro-averaged (pooled over lesions). Rand-Mean and Rand-Std summarize five iterations of random sampling. Bold marks the better value within the matched 30\% comparison.}
\label{tab:results_pat_lesion}
\renewcommand{\arraystretch}{1.2}
\setlength{\tabcolsep}{5pt}
%\resizebox{\textwidth}{!}{%
\begin{tabular}{l l | cc }
\toprule
% & & \multicolumn{3}{c|}{\textbf{Lung cancer} ($N$=35)}
%   & \multicolumn{3}{c|}{\textbf{Lymphoma} ($N$=30)}
%   & \multicolumn{3}{c|}{\textbf{Melanoma} ($N$=32)}
%   & \multicolumn{3}{c|}{\textbf{Prostate cancer} ($N$=122)}
%   & \multicolumn{3}{c}{\textbf{No cancer} ($N$=102)} \\
% \cmidrule(lr){3-5}\cmidrule(lr){6-8}\cmidrule(lr){9-11}\cmidrule(lr){12-14}\cmidrule(lr){15-17}
\begin{tabular}{l}
\textbf{Labeled} \\
\textbf{Data Budget}
\end{tabular}
&
\begin{tabular}{l}
\textbf{Sampling} \\
\textbf{Method}
\end{tabular}
  &  \begin{tabular}{cc}
       &  Lesion sens. \\
       & $\uparrow$
  \end{tabular}  & PPV  $\uparrow$ \\
\midrule
10\%  & Fixed & 0.695 & 0.801  \\
\midrule
\multirow{3}{*}{10+20\%}
  
  & Rand-Mean & 0.728 & 0.807 \\
  & Rand-Std  & 0.021 & 0.029 \\
  \cmidrule{2-4}
  & FEEDS  & \underline{0.743} &  \textbf{0.821}   \\
\midrule
100\% & Fixed  & \textbf{0.758} & 0.784 \\
\bottomrule
\end{tabular}%}
\end{table}

\subsubsection{Anatomical Region-Level Evaluation}
\begin{figure}[htbp]

  \centering
  \includegraphics[width=\linewidth]{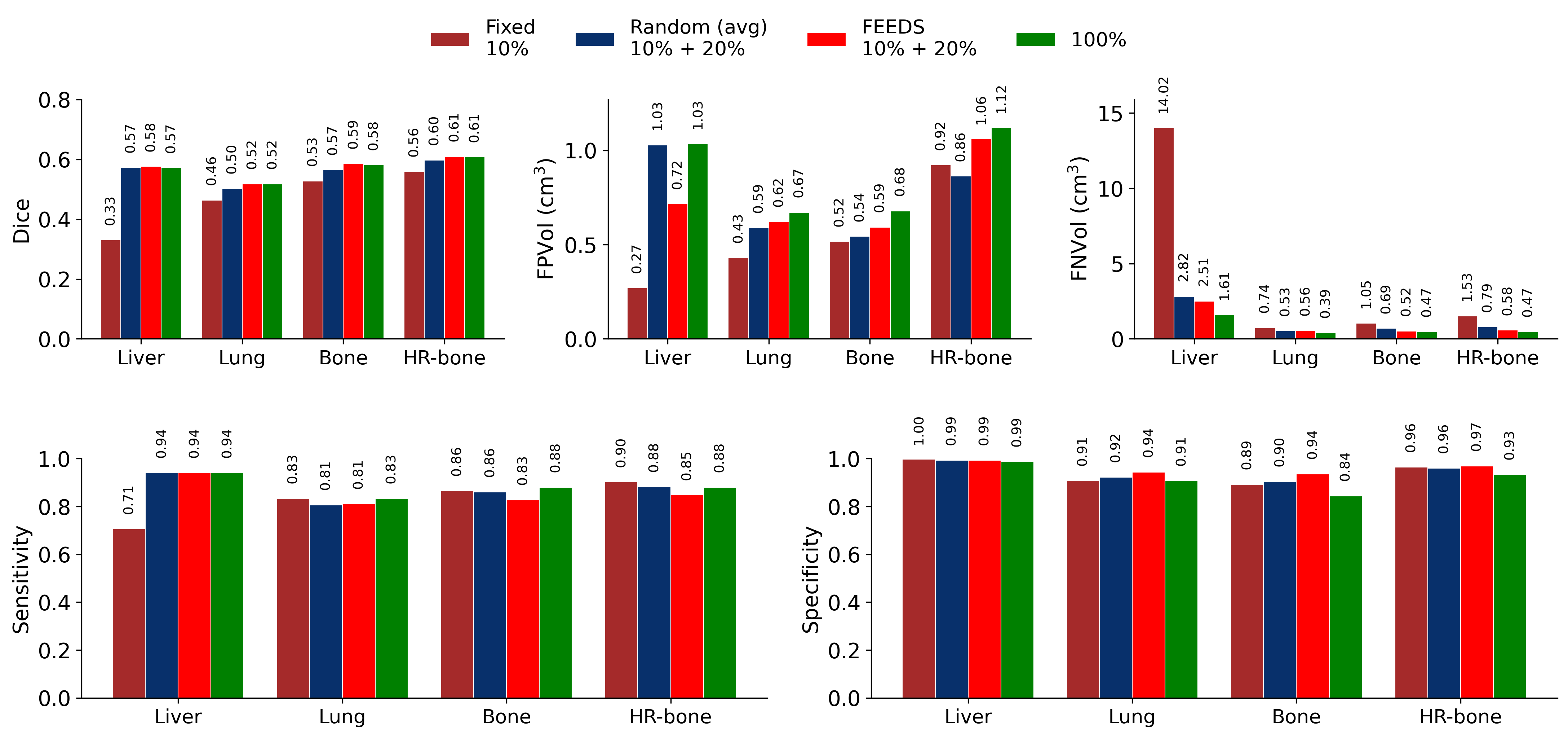}
  \caption{%
Anatomical region-level comparison of lesion segmentation with AutoPET-III test set, comparing 10\% , (10+20)\% randomly sampled (five iterations), (10 + 20)\% FEEDS-selected, and 100\% labeled training cases.
}
  \label{fig:feeds_vs_random_anatomical_bar_plots}
\end{figure}

Within the anatomic regions, lesion segmentation Dice of FEEDs is higher or same as 100\% labeled training data. FEEDS consistently yields lower FNVol than random sampling. This is clinically relevant as missed regions in these high-risk areas have high negative consequence. In the liver, FEEDS achieves lowest FPVol, suggesting least over-segmentation in this region. Sensitivity and specificity remain high and comparable across all sampling strategies, with FEEDS matching or exceeding random sampling in most regions. Notably, in bone, FEEDS achieves higher specificity than 100\% labeled training (0.94 vs.~0.84), despite using only 30\% of labeled data.

\subsubsection{Qualitative Evaluation}
\begin{figure}[htbp]
    \centering
    \begin{subfigure}[b]{0.48\textwidth}
        \centering
        \includegraphics[height=6cm]{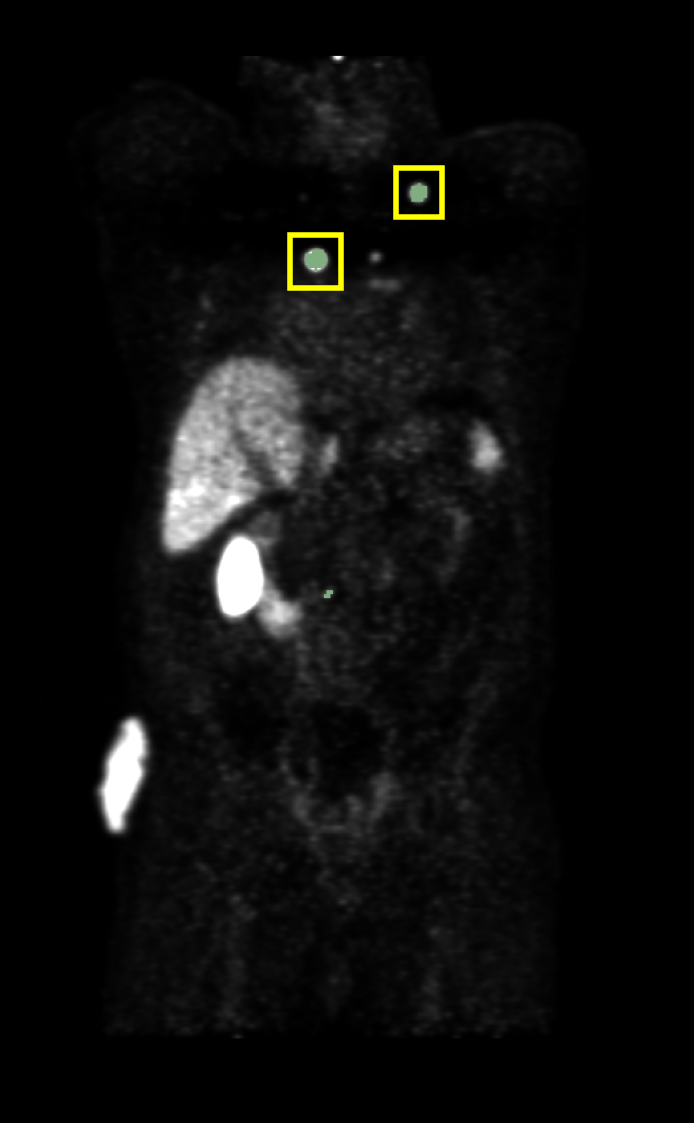}
        \caption{Example from the fixed 10\% Sample}
        \label{fig:lesions_a}
    \end{subfigure}
    \hfill
    \begin{subfigure}[b]{0.48\textwidth}
        \centering
        \includegraphics[height=6cm]{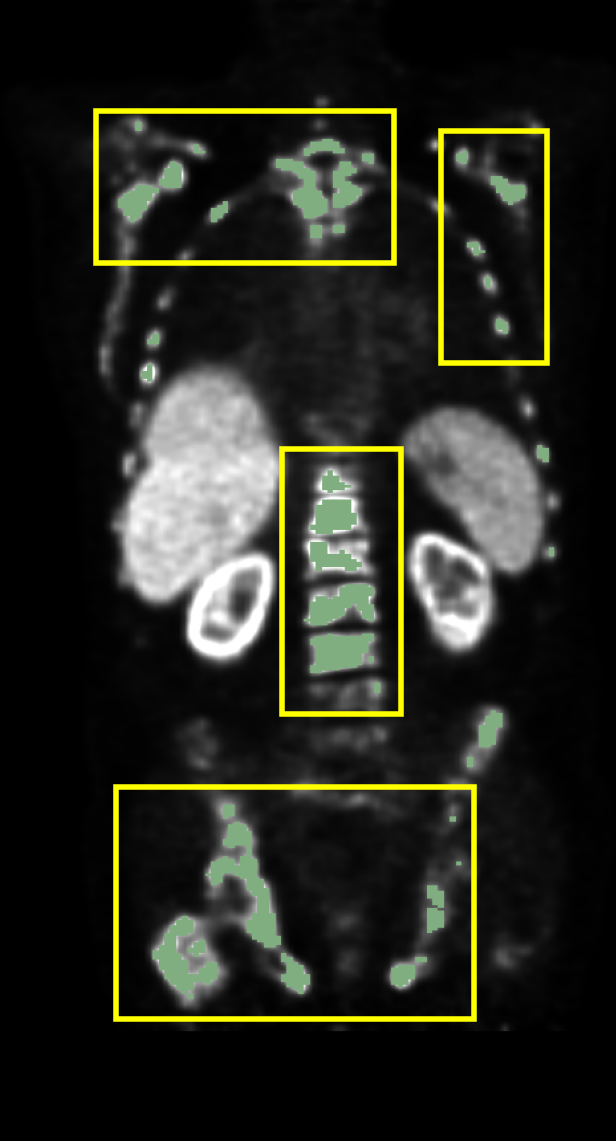}
        \caption{Corresponding farthest unlabelled case}
        \label{fig:lesions_b}
    \end{subfigure}
    \caption{Coronal view of two PSMA images with the segmented
    lesion mask (green) overlaid. Yellow boxes highlight regions of
    tracer-avid lesion. (a) shows only a small
    number of lesions, whereas (b) shows significant tumor burden }
    \label{fig:lesion_comparison}
\end{figure}

\textbf{FEEDS Interpretability:}
Pairwise comparison of closest and farthest cases in the DinoV2 embedding space showed that the closest scans were often longitudinal scans of the same patient, or scans from two patients with very similar disease distributions. Farthest scans within a tracer type show wide variability in the location and extent of cancer spread, as demonstrated in an example PSMA-pair in Figure \ref{fig:lesion_comparison}. Thus, FEEDS enables adding diverse cases to training samples.

\begin{figure}[htbp]
  \centering
  \includegraphics[width=\linewidth]{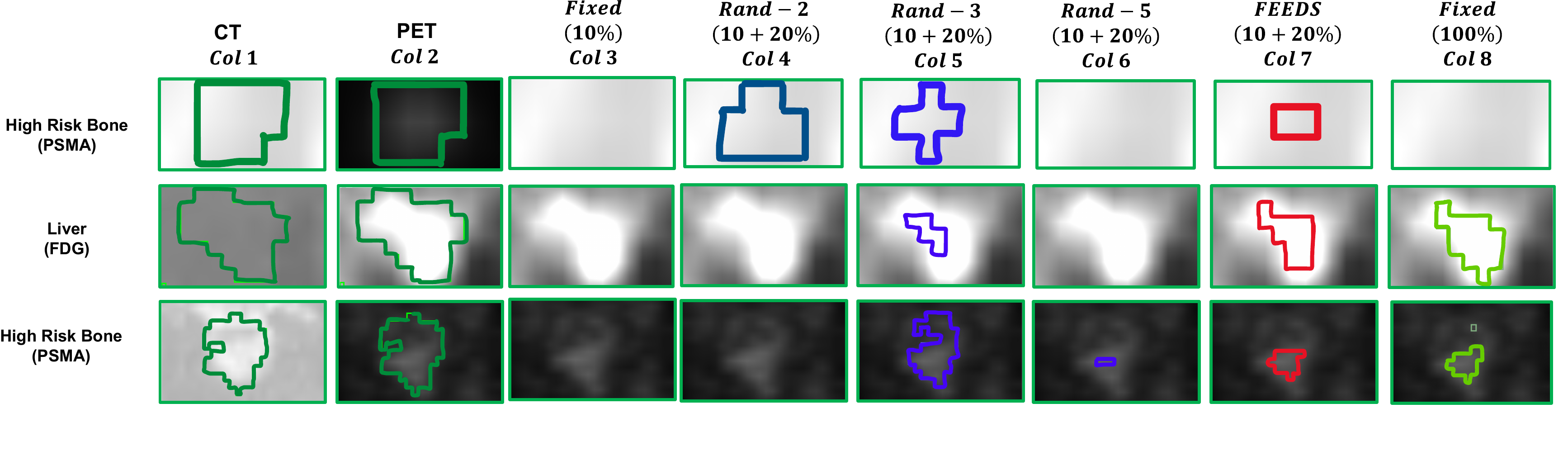}
  \caption{%
Qualitative evaluation across three representative lesion regions. Columns 1 and 2 show the CT and PET patches, respectively. Column 3 shows the prediction using 10\% labeled data, Columns 4–6 use an additional 20\% randomly selected cases, Column 7 shows FEEDS, and the final column shows the fully supervised 100\% upper bound. Rows 1 and 3 show PSMA-avid high-risk bone lesions, while Row 2 shows an FDG-avid liver lesion. Unlike the 10\% and random (10+20)\% models, FEEDS successfully detects the missed lesions.
}\label{fig:qual_results}
\end{figure}

\begin{figure}[htbp]
  \centering
  \includegraphics[width=\linewidth]{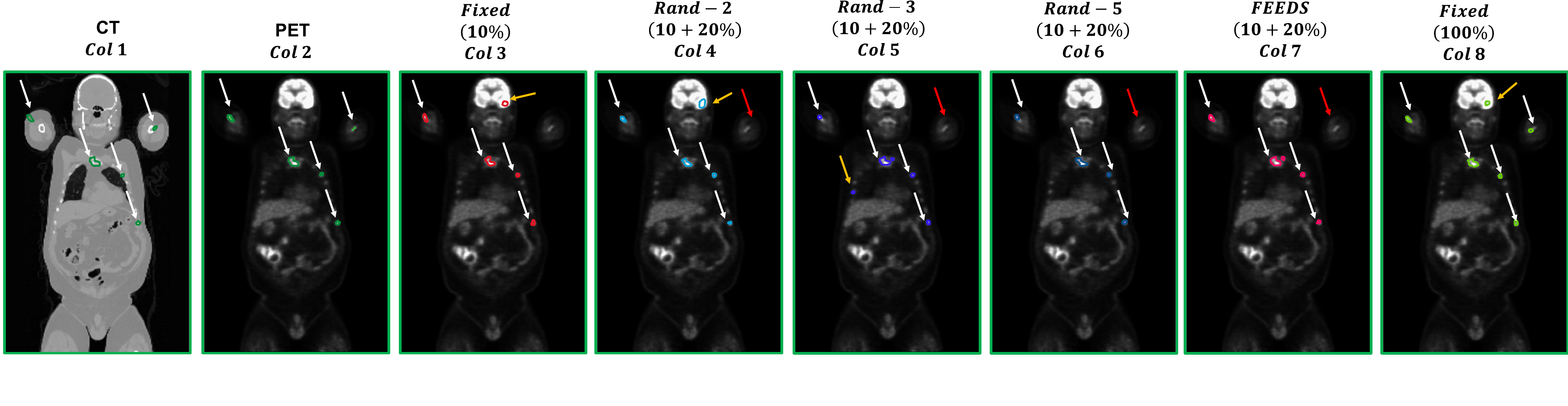}
  \caption{%
Qualitative evaluation of FEEDS on a whole-body PET/CT scan. Columns 1 and 2 show the CT and PET images, respectively, with the reference lesions indicated by white arrows. We also identify the false-positive lesions with orange arrows and false negative lesions with red arrows.  Column 3 presents a false-positive prediction in the skull from the fixed model trained with 10$\%$ labeled data. Columns 4–6 show predictions from three models trained using the same fixed $10\%$ with randomly selected $20\%$ of the labeled data. Rand-2 in Column 4 also produces a false positive in the skull, whereas Rand-3 in Column 5 produces a false positive near the right lung. Rand-5 in Column 6 and the proposed FEEDS method in Column 7 detect the lesions without additional false-positive predictions. Column 8 shows the fully supervised model trained with $100\%$ of the labeled data, which produces a similar false positive in the skull.
}\label{fig:qual_results_wb}
\end{figure}

\textbf{Qualitative comparison with random sampling:} FEEDS is robust to performance variability from random sampling stochasticity, detects more lesions than models trained with randomly-sampled training data at the same labeled data budget, has fewer false positives, shows improvements across all anatomical regions, disease types, and tracers, and achieves performance similar to models trained with 100\% labeled data at 70\% reduced annotation burden (Figures \ref{fig:qual_results}, \ref{fig:qual_results_wb}).

\subsection{Performance on Unseen Datasets (Table \ref{tab:results_pat_lesion_unseen})}
On Deep-PSMA, FEEDS achieves the lowest FPVol and outperforms random sampling in FNVol (14.97 vs.~17.07) and lesion sensitivity (0.658 vs.~0.624), approaching the fully supervised model. On DH, the model trained with the 10\% labeled training data achieves the highest Dice and lowest FPVol, with FEEDS being second-highest Dice (0.536 vs.~0.534) and second-lowest FPVol (16.034 vs.~19.081). FEEDS achieves the best FNvol and the second-best lesion sensitivity, closely following 100\% labeled training. Interestingly, unlike the other datasets, the fully supervised AutoPET model produces more false positives and a lower Dice score on DH. FEEDS achieves a balance and generalizable performance on both unseen datasets. 

\begin{table}[!htbp]
\centering
\caption{Voxel-level and Lesion-level evaluation on the unseen Deep-PSMA and DH dataset to assess generalizability. Lesion sensitivity and PPV are reported micro-averaged (pooled over lesions). Rand-Mean and Rand-Std summarize five iterations of random sampling. Within each dataset, \textbf{bold} indicates the best value and \underline{underline} the second-best value among Fixed 10\%, Rand-Mean, FEEDS, and Fixed 100\% (Rand-Std is excluded, as it is a variability measure rather than a performance score).}
\label{tab:results_pat_lesion_unseen}
\small
\renewcommand{\arraystretch}{1.2}
\setlength{\tabcolsep}{3.5pt}
%\resizebox{\textwidth}{!}{
\begin{tabular}{l c c| ccc  cc}
\toprule
\textbf{Dataset} & \makecell{\textbf{Labeled Data }\\\textbf{Budget}}
&
\begin{tabular}{l}
\textbf{Sampling} \\
\textbf{Method}
\end{tabular} & \multicolumn{3}{c}{\textbf{Voxel-level}}
& \multicolumn{2}{c}{\textbf{Lesion-level}}
\\
\cmidrule(lr){4-6}
\cmidrule(lr){7-8}
 & & & Dice $\uparrow$ & FP Vol. $\downarrow$  & FN vol. $\downarrow$ & \begin{tabular}{cc}
       &  Lesion sens. \\
       &  (Micro) $\uparrow$
  \end{tabular}   & Lesion PPV $\uparrow$    \\ 
\midrule
\multirow{5}{*}{\makecell{\textbf{Deep-PSMA }\\\textbf{($N=200$)}}} & 10\%  & Fixed & 0.647 & \underline{4.380} & 20.352  & 0.580 & \underline{0.912}  \\
\cmidrule{2-8}
& \multirow{3}{*}{10+20\%}
 & Rand-Mean & \underline{0.653} & 5.556 & 17.069 & 0.624 & 0.912  \\
 & & Rand-Std  & 0.008 & 1.464 & 2.297  & 0.035 & 0.014 \\
  \cmidrule{3-8}
  & &  FEEDS  & 0.645 & \textbf{3.389} & \underline{14.979}    &  \underline{0.658} &  \textbf{0.921} \\ 
\cmidrule{2-8}
& 100\% & Fixed & \textbf{0.655} & 7.520 & \textbf{10.63} & \textbf{0.686} & 0.892 \\
\midrule[1.2pt]
\multirow{5}{*}{\makecell{\textbf{DH }\\\textbf{($N=23$)}}} & 10\%  & Fixed & \textbf{0.536} & \textbf{16.034} & 3.084  & 0.749 & \textbf{0.449}  \\
\cmidrule{2-8}
& \multirow{3}{*}{10+20\%}
 & Rand-Mean & 0.491 & 23.282 & 2.889 & \underline{0.763} & 0.437 \\
 & & Rand-Std  & 0.017 & 4.781  & 0.258 & 0.036 & 0.018 \\
  \cmidrule{3-8}
  & &  FEEDS  & \underline{0.534} & \underline{19.081} & \textbf{2.721}    &  \underline{0.766} & \underline{0.446}  \\
\cmidrule{2-8}
& 100\% & Fixed &  0.489 & 34.728 & \underline{2.883} & \textbf{0.774} & 0.382  \\
\bottomrule
\end{tabular}
\end{table}

\subsection{Statistical Analysis}

We evaluate statistical significance within and across all Datasets, comparing FEEDS against Random using a two-sided Wilcoxon signed-rank test with FDR-BH correction $(\alpha = 0.05)$. Non-inferiority was assessed with a one-sided paired t-test $(\alpha = 0.05)$ on the per case difference with the 100\% model. Our non inferiority margins were $0.05$ for Dice, and $5 \text{ cc}$ for false-positive and false-negative volumes. Each metric was considered as a separate family of tests across 
all of the datasets. 

FEEDS significantly reduces false-positive volume when compared with random sampling $(p < 0.01)$ in all datasets, while also remaining non inferior in Dice, false-negative, and false-positive volumes against the 100\% model in our pooled analysis. Furthermore, in the DH cohort, FEEDS showed statistically significant improvement in all three metrics when compared with random sampling. 
These results are summarised in Table \ref{tab:pooled_statistical_significance}.  

% FEEDS also consistently reduces false negatives  relative to random sampling, though these differences do not reach statistical significance after multiple comparison correction. 

% For all test cases, false negative volume was lower for FEEDS but was not statistically significant (p = 0.6). FEEDS matches random sampling on Dice while staying non-inferior to the model trained with 100\% labeled data budget. 

\begin{table}[t]
\centering
\caption{Pooled results comparing FEEDS against Random sampling across datasets.
Dice, false positive (FP) volume, and false negative (FN) volume are reported.
FEEDS mean and Random mean are shown, with the mean difference
$\Delta$ (FEEDS~$-$~Random). Statistical significance is assessed with a two-sided
Wilcoxon signed-rank test (FDR-BH corrected $p$, $\alpha = 0.05$).
Non-inferiority (NI) against the 100\% model is assessed with a one-sided paired
$t$-test ($\alpha = 0.05$; margins: 0.02 Dice, 2\,mL FP/FN volume);
\checkmark{} indicates NI passed.}
\label{tab:pooled_statistical_significance}
\begin{tabular}{@{}llcccccc@{}}
\toprule
Dataset & Metric & $n$ & FEEDS & Random & $\Delta$ & $p_{\text{Wilcoxon}}$ & NI \\
 &  &  & (mean) & (mean) & (mean) &  &  \\
\midrule
\multirow{3}{*}{AutoPET}  & Dice   & 207 & 0.613  & 0.613  & $-0.001$ & $0.325$        & \checkmark \\
                          & FP Vol & 322 & 6.59   & 7.72   & $-1.12$  & \num{2.20e-26} & \checkmark \\
                          & FN Vol & 207 & 11.20  & 12.27  & $-1.07$  & $0.760$        & \checkmark \\
\midrule
\multirow{3}{*}{DeepPSMA} & Dice   & 200 & 0.645  & 0.653  & $-0.009$ & $0.447$        & \checkmark \\
                          & FP Vol & 200 & 3.39   & 5.56   & $-2.17$  & \num{6.75e-18} & \checkmark \\
                          & FN Vol & 200 & 14.98  & 17.07  & $-2.09$  & $0.242$        & $\times$ \\
\midrule
\multirow{3}{*}{DH}       & Dice   & 23  & 0.534  & 0.491  & $+0.043$ & $0.050$        & \checkmark \\
                          & FP Vol & 23  & 19.08  & 23.28  & $-4.20$  & $0.028$        & \checkmark \\
                          & FN Vol & 23  & 2.72   & 2.89   & $-0.17$  & $0.948$        & \checkmark \\
\midrule
\multirow{3}{*}{All}      & Dice   & 430 & 0.623  & 0.625  & $-0.002$ & $0.411$        & \checkmark \\
                          & FP Vol & 544 & 5.94   & 7.58   & $-1.64$  & \num{5.57e-44} & \checkmark \\
                          & FN Vol & 430 & 12.51  & 14.00  & $-1.49$  & $0.603$        & \checkmark \\
\bottomrule
\end{tabular}
\end{table}
\section{Discussion}
\label{sec:discussion}

We present FEEDS (Foundation Model-Enabled Efficient Data Sampling), a label- and compute-efficient training method for lesion segmentation in pan-cancer, multi-tracer whole-body PET/CT. FEEDS selects diverse, representative training samples using foundation model features, which are then annotated and added to the labeled set for supervised training. Unlike unsupervised, semi-supervised, and active learning strategies, it requires no iterative training or large compute resources. We trained and validated FEEDS on AutoPET-III, comparing with random sampling, Determinantal Point Processes (DPP), and pseudolabel-guided semi-supervised learning. We evaluated FEEDS on a held-out AutoPET-III test set plus two unseen datasets (Deep-PSMA and DH) using voxel-, lesion-, anatomic region-, and disease/tracer-stratified metrics. FEEDS outperformed random sampling and generalized well, matching 100\% labeled-data performance with only a 30\% labeling budget. It improved lesion sensitivity, PPV, and false negative volume across diseases, tracers, and anatomic regions, while maintaining Dice and false positive volume even at low labeling budgets, supporting its clinical relevance.

The farthest-first selection in DINOv2 embedding space works by targeting gaps in the current training distribution. Pan-cancer PET datasets contain highly variable lesion burden, disease characteristics, tracer kinetics, and anatomy. Selecting maximally dissimilar cases exposes the model to this diversity, improving generalization over random sampling, which may over-represent common patterns, particulary in small, imbalanced, labeled datasets.

Unlike prior PET/CT segmentation studies that rely solely on voxel-level evaluation, we provide lesion- and anatomic region-level assessments relevant to treatment planning. Since cancer location, extent, and type affect both prognosis and treatment decisions, models should be evaluated on their ability to detect high-risk lesions specifically (e.g., liver, lung, high-risk bone lesions), not just voxel-level segmentation in aggregate. To our knowledge, no prior work has assessed auto-segmentation performance from this treatment-planning perspective.

A notable finding was that the model trained with 100\%-labeled AutoPET-III training data showed lower Dice in the DH dataset, than the 10\%-budget model. This is contrary to trends in the AutoPET-III and Deep-PSMA test sets. We attribute this to scanner and imaging protocol differences across datasets. AutoPET-III PSMA scans came from three scanners (GE Discovery 690; Siemens Biograph 20 mCT Flow 20 and 64-4R TruePoint), with FDG scans from a Siemens Biograph mCT. Deep-PSMA included GE Discovery 710/690, Siemens Biograph, and Siemens Vision 600 scanners. DH used a Siemens Biograph Vision 600 scanner. The substantial technological gap between the older AutoPET-III scanners and the newer, fully digital Vision 600 in DH, likely limited generalization of AutoPET-III-trained models. Notably, FEEDS at 30\% budget achieved higher Dice and lower false positive and false negative volumes than the fully trained AutoPET-III model.

Our study has several limitations. First, FEEDS uses DINOv2 features from 2D PET maximum intensity projections rather than 3D volumes, which may lose spatial information. Future work will explore 3D-native foundation models. Second, DINOv2 is a general-purpose model trained on natural images \cite{oquab2023dinov2}, not radiology- or PET-specific, though it has shown strong performance on medical images \cite{baharoon2023evaluating, bhattacharya2025aggressiveness}. We found it captured training set diversity well, but we will compare with other domain-specific foundation models in the future. Third, we used only PET features for diversity selection, and incorporating CT features alongside PET is a direction for future work.

FEEDS is designed for practical deployment. Clinical archives often contain large volumes of unannotated PET/CT scans, and manual annotation is labor-intensive for already overburdened radiologists. FEEDS can rank unlabeled cases by distance from a small labeled set in foundation-model feature space, producing a prioritized annotation queue so radiologists focus only on the most informative cases. Beyond one-time training, FEEDS can support few-shot learning as new unlabeled data becomes available, selecting the most diverse cases for annotation. Because DINOv2 is deployment-friendly, feature extraction is straightforward. The workflow is tracer-agnostic, requires no task-specific training for sample selection, and scales efficiently to large unannotated repositories.

\section{Conclusion}
\label{sec:conclusion}

We presented FEEDS, a foundation model-enabled sampling strategy for label- and compute-efficient training of lesion segmentation models on large, unlabeled oncological PET datasets. Using DINOv2 embeddings to select diverse, representative cases for annotation, FEEDS outperforms random selection, pseudo-label semi-supervised learning, and limited-data baselines across pan-cancer, multi-tracer segmentation tasks. It matches fully-labeled (100\%) performance with far fewer annotations and generalizes well across three independent test sets. Being tracer- and cancer-type agnostic with no need for feature extractor adaptation, FEEDS is broadly applicable to clinical annotation prioritization pipelines.
% ============================================================
\section*{Acknowledgement}
We gratefully acknowledge support from the Munck-Pfefferkorn Research Funds from the Geisel School of Medicine at Dartmouth College, the American Cancer Society Institutional Research Grant, and support from the Department of Biomedical Data Science at Dartmouth College. Research reported in this publication was also supported by an Institutional Development Award (IDeA) from the National Institute of General Medical Sciences of the National Institutes of Health under grant number 1P30GM149408. All of this support was instrumental in making this work possible.

\section*{Author Contributions}

All authors contributed to the conception and design of the study. Material preparation, data collection, and data analysis were performed by Biratal Raj Wagle, Bashirul Azam Biswas, and Grant Chau. Clinical expertise and interpretation were provided by Matthew E. Maeder, Muhammad Azeem Arshad, James Yu, and Michael S. Leapman. Lesion annotations for the Dartmouth Hitchcock Medical Center cohort were performed by Bashirul Azam Biswas and clinically reviewed by James B. Yu. Indrani Bhattacharya supervised all stages of the study, funding acquisition, mentorship, technical guidance, and manuscript preparation. The initial draft of the manuscript was written by Biratal Raj Wagle, and all authors reviewed and commented on subsequent versions. All authors read and approved the final manuscript.

% ============================================================
\bibliographystyle{plain}
\bibliography{sample}

\end{document}